# Machine Learning-Based Genomic Linguistic Analysis (Gene Sequence Feature Learning): A Case Study on Predicting Heavy Metal Response Genes in Rice

Ruiqi Yang[a, b*], Jianxu Wang[a], Wei Yuan[a], Xun Wang[a], Mei Li[b]

[a]*State Key Laboratory of Environmental Geochemistry, Institute of Geochemistry, Chinese Academy of Sciences, Guiyang, Guizhou 550082, P. R. China*

[b]*Northwest A&F University, Yangling, Shaanxi 712100, P. R. China*

---

[*] Corresponding author.
Email addresses: rickyyoung@nwafu.edu.cn (Ruiqi Yang).

## Abstract

This study explores the application of machine learning-based genetic linguistics for identifying heavy metal response genes in rice (Oryza sativa). By integrating convolutional neural networks and random forest algorithms, we developed a hybrid model capable of extracting and learning meaningful features from gene sequences, such as k-mer frequencies and physicochemical properties. The model was trained and tested on datasets of genes, achieving high predictive performance (precision: 0.89, F1-score: 0.82).

RNA-seq and qRT-PCR experiments conducted on rice leaves which exposed to $Hg^0$, revealed differential expression of genes associated with heavy metal responses, which validated the model's predictions. Co-expression network analysis identified 103 related genes, and a literature review indicated that these genes are highly likely to be involved in heavy metal-related biological processes. By integrating and comparing the analysis results with those of differentially expressed genes (DEGs), the validity of the new machine learning method was further demonstrated.

This study highlights the efficacy of combining machine learning with genetic linguistics for large-scale gene prediction. It demonstrates a cost-effective and efficient approach for uncovering molecular mechanisms underlying heavy metal responses, with potential applications in developing stress-tolerant crop varieties.

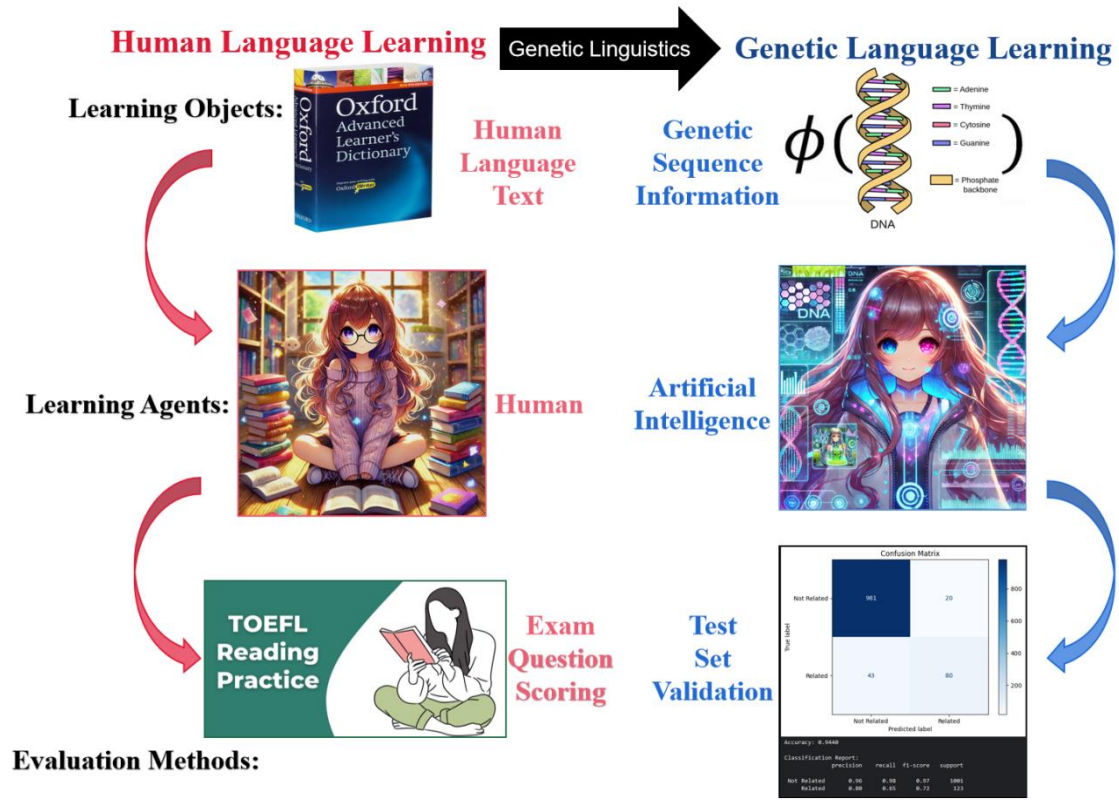

## Synopsis

High-throughput prediction of rice heavy metal response genes was achieved using a convolutional neural network-random forest hybrid model based on the concept of genomic linguistics.

## Introduction

The harm caused by heavy metals in rice refers to the absorption of heavy metal elements such as cadmium, lead, and mercury from soil or water during rice growth, leading to excessive levels of heavy metals within the plant. This negatively impacts the growth, yield, and quality of the rice [1]. Long-term consumption of rice with excessive heavy metal levels can result in health issues such as kidney damage [2], neurological disorders [3], and even cancer [4]. To mitigate these risks, it is crucial to study and develop rice varieties that can specifically respond to heavy metal contamination in designated environments. However, current genetic research on rice remains incomplete, and the annotated information in rice gene databases is very limited. Traditional biological experiments are costly, time-consuming, and inefficient, while conventional bioinformatics methods, primarily reliant on sequence alignment, struggle to uncover the deeper meanings within gene sequences.

Currently, Natural Language Processing (NLP) technology has made significant progress, with its core goal being to enable computers to understand, generate, and process human language. Based on the principles of NLP, artificial intelligence systems can achieve a deep understanding of human language through semantic analysis, syntactic parsing, and contextual comprehension [5]. For example, applications such as machine translation [6], voice assistants [7], and intelligent customer service [8] have been widely integrated into daily life, demonstrating the practical achievements of NLP technology. Despite challenges such as contextual ambiguity and cultural differences, the rapid development of NLP has brought artificial intelligence closer to or even beyond human-level language understanding, laying a solid foundation for human-computer interaction and intelligent applications.

Just as the English language is composed of letters such as A, B, C, and D, gene sequences are made up of A, G, C, and T. Current research suggests that gene sequences exhibit certain regularities, such as conserved sequences that can perform identical or similar functions across different species [9]. As shown in Figure 1, gene sequences can also be viewed as a form of biological language [10]. Using machine learning to decode the implicit information within gene sequences not only enhances the efficiency of studying heavy metal response genes but also deepens our understanding of how rice genes influence the plant's response to heavy metals.

**Figure 1**

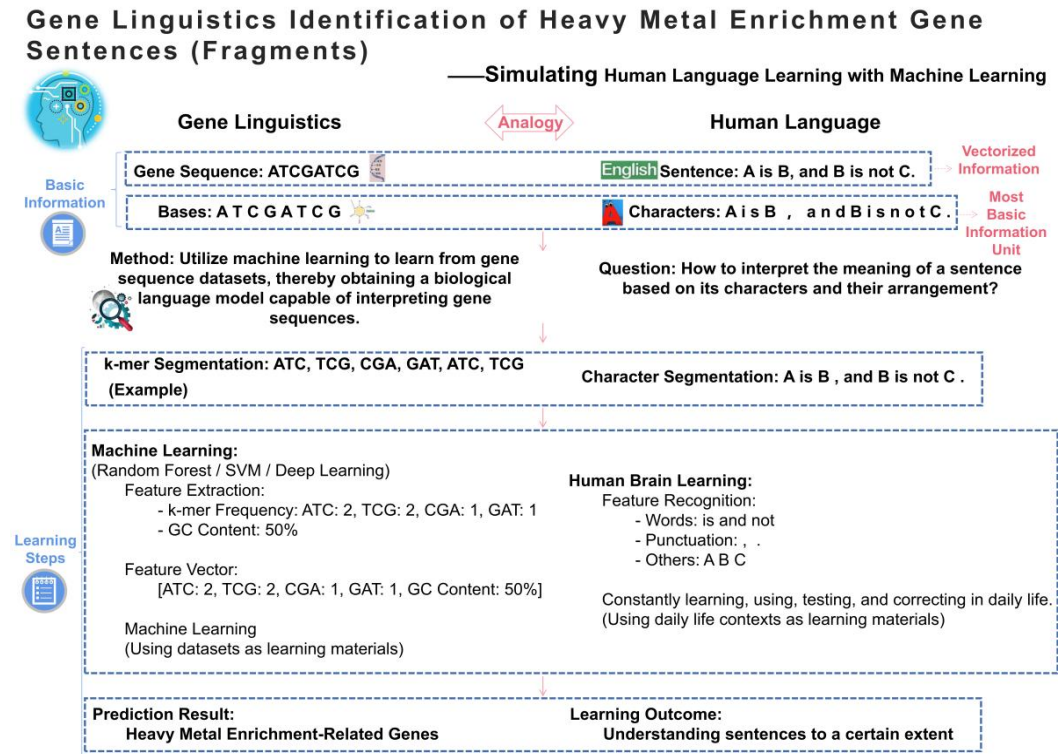

Figure 1. Application of machine learning-based genetic linguistics in identifying heavy metal accumulation genes.

Figure 1 illustrates how genomic linguistics can be used to identify genes related to heavy metal accumulation. By drawing an analogy to the process of learning human languages, machine learning methods are employed to extract features from gene sequence data (such as k-mer frequency [11] and GC content [12]) and to build a biological language model to interpret gene sequences. Through iterative learning, testing, and refinement, the model is able to recognize and understand genes associated with heavy metal accumulation, showcasing the potential of machine learning in gene analysis.

## Materials and Methods

**Figure 2**

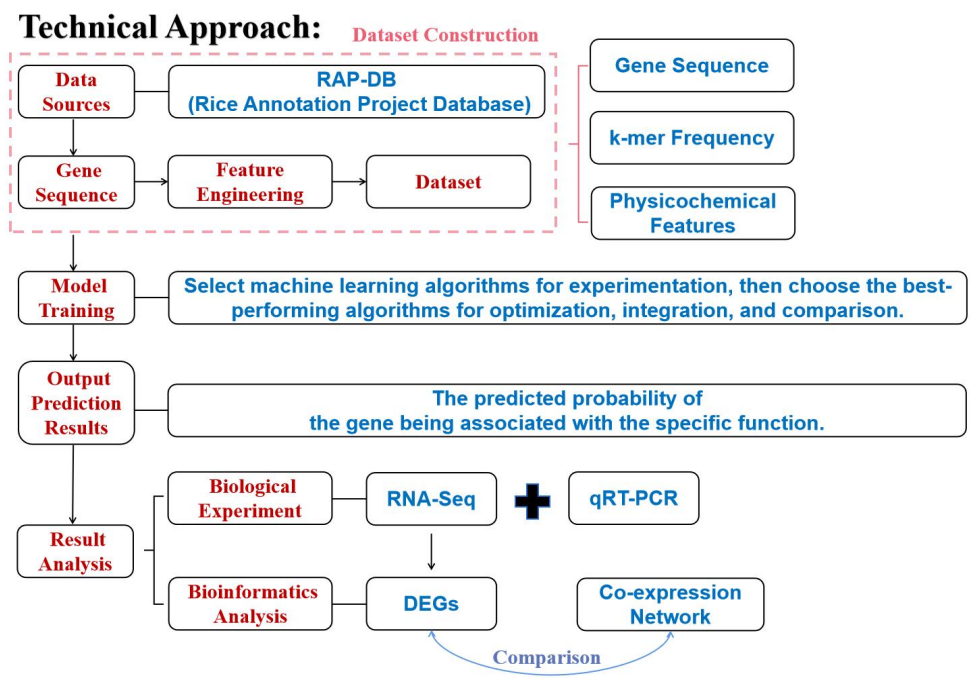

Figure 2. Schematic diagram of the technical approach in this study.

**Collection of Heavy Metal-Related Gene Sequence Information in Rice**

By searching for "heavy metal" in the RAP-DB database, over a hundred heavy metal-related genes were identified. The complete sequences, coding sequences, and various transcript sequences for the same gene were separately collected to expand the dataset, ultimately resulting in nearly 400 heavy metal-related sequence data entries. To ensure randomness in non-related sample sequences, the term "rice" was searched and similarly expanded, generating more than 3,000 non-related sample sequences. The total dataset of related and non-related sequences comprised nearly 4,000 entries, with a ratio of approximately 1:10. Detailed information can be found in Text S1 and Csv S1 (Training Set Samples). To ensure randomness in prediction samples, the letter "b" was used for search, collecting random prediction samples as shown in Zip S1 (Prediction Set Samples).

**Characterizing Gene Sequences into Digital Vectors**

The raw data files include both xlsx and fasta files, which were processed into csv files using Biopython packages, as demonstrated in the corresponding code files in Zip S2 (Code). The detailed process is described in Text S2. Subsequently, concepts from NLP text feature extraction were applied, such as the Bag-of-Words model, TF-IDF (Term Frequency-Inverse Document Frequency), Word2Vec, and One-Hot Encoding. Through feature engineering, the gene sequences were transformed into feature vectors containing information such as sequence length, GC content, physicochemical properties, and k-mers. Refer to Text S2, Figure S1, and Figure S2 for details.

**Machine Learning Models for Gene Prediction**

Using the Kaggle platform (www.kaggle.com) for code execution, various machine learning models were applied to the dataset constructed from the search keywords "heavy metal" and "rice" for training and validation. After experimentation, optimization, and comparison of algorithms, the convolutional neural network-random forest hybrid model was ultimately selected. Refer to Text S3, Figure S3, and Zip S2 (Code) for details. The dataset constructed from the search keywords "heavy metal" and "rice" was used as specialized training material for comprehensive model training, while the dataset constructed from the search letter "b" was used as the prediction set. This yielded predictions on whether the respective rice samples exhibit heavy metal responses, with the output results provided in Csv S2 (Output Results).

**Biological Experiments and Bioinformatics Analysis for Auxiliary Validation**

RNA extraction and transcriptome sequencing (RNA-Seq) were used to study gene expression changes in rice. The reliability of the RNA-Seq results was validated using real-time quantitative PCR (qRT-PCR). Through RNA-Seq data, differentially expressed genes (DEGs) in rice leaves under $Hg^0$ exposure were identified, including the upregulation and downregulation of various genes at different time points (3, 9, and 24 hours), as shown in Text S4 and Xlsx S1.

Samples with prediction probabilities greater than 0.7 in Csv S2 (Output Results) were selected and analyzed using Co-Expression Network (CoExp Network) analysis, as detailed in Text S5 and Xlsx S2 (CoExp Network Node List). By integrating this information with data from Xlsx S1, the results were further organized into Xlsx S3 and Xlsx S4. The findings demonstrate the rationality and interpretability of the machine learning results, which exhibit stronger information mining and predictive capabilities compared to traditional bioinformatics analyses. This advancement marks a transition from traditional sequence alignment-based bioinformatics to artificial intelligence models with gene sequence learning and understanding capabilities.

## Results And Discussion

**Characteristics of the Dataset**

The feature importance analysis of the random forest model indicates no significant differences among various physicochemical properties, as shown in Figure 3 (a). This is because the physicochemical values extracted during feature engineering essentially reflect the mapping of the proportions of A, G, C, and T in the sequences rather than their actual physicochemical properties.

The analysis of k-mer sequences using the random forest model reveals significant differences in their impact. For instance, among the top 20 most important 3-mer sequences shown in Figure 3 (b), the top-ranked sequence, TTT, has an importance score exceeding 0.05, while the 20th-ranked sequence, CGT, has a score below 0.02. The impact of the former is approximately 2.5 times greater than that of the latter.

**Figure 3**

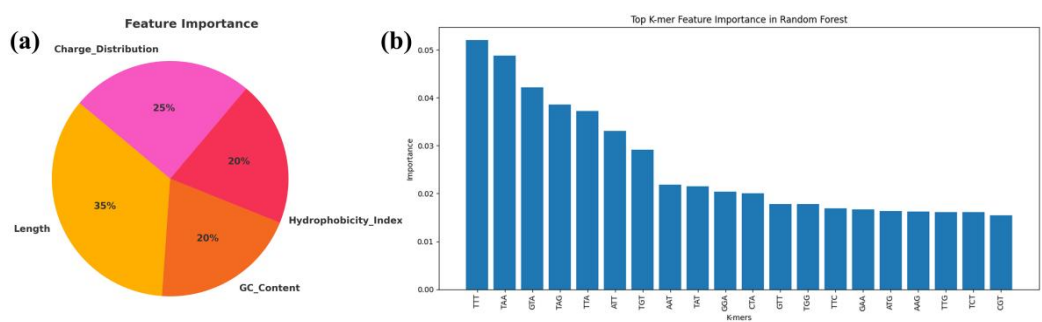

Figure 3. (a) shows the contribution proportions of overall features (e.g., length, charge distribution, hydrophobicity index, and GC content) to the model's importance, (b) displays the importance ranking of different k-mers (short sequence fragments) in the random forest model.

This indicates that the differences among features in the dataset are determined by the underlying relationships within the sequences. Essentially, all features in the dataset are mappings of sequence information. The machine learning process of feature learning is fundamentally a process of learning the sequences themselves—a process of learning and understanding the "language of life" encoded in gene sequences.

**Machine Learning Performance**

To compare the learning performance of different algorithms, each algorithm was run five times, and the best performance for predicting heavy metal-related genes was recorded in Table S1. Since the aim of this study is to improve the efficiency of gene research, "precision" was chosen as the primary metric to ensure positive results are as accurate as possible, thereby reducing the experimental costs of future biological validations.

As shown in Table S1, simple, unmodified models outperformed more complex models, likely because the limited annotations currently available in the database are insufficient to support the complexity of advanced models. Some models performed poorly during training, indicating their inability to understand the existing gene sequence features.

Ultimately, the results showed that among traditional machine learning algorithms, the unmodified random forest, and among deep learning algorithms, the unmodified convolutional neural network (CNN), achieved the best training performance. Combining the two into a CNN-random forest hybrid model further enhanced performance significantly. Moreover, compared to other deep learning algorithms, this hybrid model greatly reduced runtime during practical execution, resulting in a substantial improvement in model efficiency.

In terms of prediction performance, the newly developed convolutional neural network-random forest hybrid model demonstrates satisfactory classification effectiveness, as shown in Figure 4 (a).

Regarding model stability, the performance metrics (Precision, Recall, and F1-score) of the CNN-random forest hybrid model show minimal variation across different experimental iterations, with a relatively stable trend, as illustrated in Figure 4 (b). This indicates that the CNN-random forest hybrid model exhibits excellent stability in understanding gene sequences.

**Figure 4**

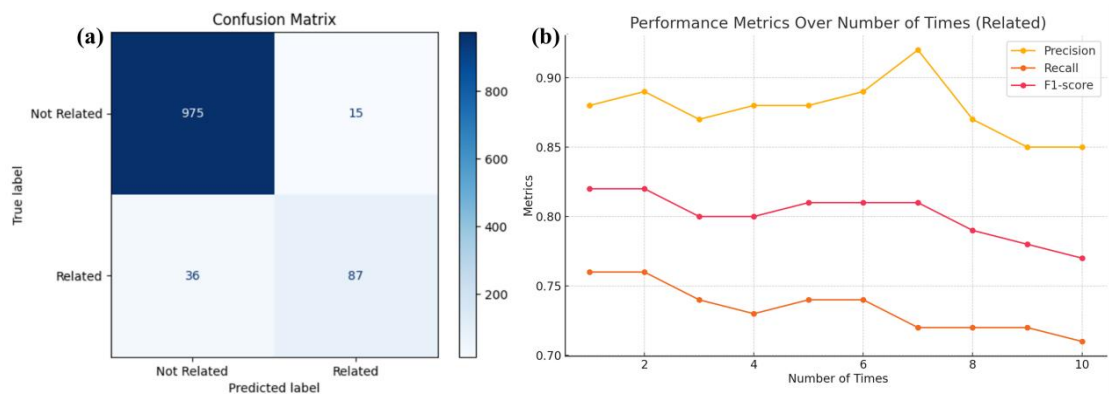

Figure 4. (a) displays the confusion matrix, showing the model's prediction accuracy for the "Related" and "Not Related" categories, including classification results such as true positives and false positives, (b) illustrates the performance metrics (Precision, Recall, and F1-score) trends across different experimental iterations, evaluating the model's overall stability and performance.

Random forest and convolutional neural networks (CNN) demonstrate strong learning performance on gene sequences, whereas other algorithms perform poorly, likely due to various factors.

In terms of data characteristics and model adaptability, random forest and CNN exhibit strong versatility. Random forest, as a non-parametric machine learning algorithm, can handle high-dimensional data and nonlinear feature relationships, making it particularly suitable for complex feature data like gene sequences [13]. CNN excels at extracting local features and patterns, making it ideal for analyzing spatial or adjacent relationships in gene sequences [14].

Conversely, other algorithms may not align well with the characteristics of gene sequence data. Graph neural networks (GNN) [15] and long short-term memory networks (LSTM) [16] have high requirements for data volume and annotation quality, which gene data may lack in terms of rich annotations or long-sequence dependency characteristics. Transformer models excel at handling large-scale data but may be less optimized for data like gene sequences, which primarily exhibit local features [17]. The Naive Bayes algorithm, with its strong assumption of data independence, struggles to capture the dependency relationships inherent in gene sequences [18].

Regarding data scale and quality, gene sequence data is often high-dimensional but limited in volume, with potentially insufficient annotations. This can hinder the training of complex models (e.g., Transformer or LSTM), leading to overfitting or suboptimal performance. Certain algorithms, such as SMOTE and other oversampling methods [19], may introduce noise when expanding the dataset, reducing model accuracy.

In terms of model complexity and generalization ability, complex models (e.g., Transformer or CNN with attention mechanisms) typically require a large amount of data to support their training. When data is limited or lacks diversity, simpler models (e.g., basic CNN or random forest) with stronger generalization abilities may outperform them and are better suited for current learning tasks with limited data.

Concerning the impact of target metrics, this study prioritizes precision [20] as the primary evaluation metric rather than a comprehensive metric like F1-score [21]. Many algorithms (e.g., SVM [22] or CatBoost [23]) may favor improving recall [24] at the expense of precision, leading to lower performance in this evaluation.

In terms of the ability to understand features, random forest, through its random feature selection and ensemble of decision trees, can handle feature correlations and complex nonlinear relationships [25], which is particularly important for gene sequence data. Convolutional neural networks (CNN) can automatically extract important features from data without the need for manual feature engineering [26], making them especially suitable for the complex positional and sequential correlations in gene sequences. In contrast, other algorithms may struggle to effectively extract the key feature patterns within gene sequences.

In terms of experimental design and parameter optimization, simple machine learning algorithms (e.g., random forest) perform well without complex modifications, likely because their default settings are well-suited to the data. In contrast, complex models require extensive parameter tuning (e.g., learning rate, number of layers, feature extraction methods) to fully realize their potential; otherwise, their performance may be suboptimal.

Overall, the superior performance of random forest and convolutional neural networks (CNN) in gene sequence research can be attributed to their strong adaptability to data characteristics, relatively low dependency on large datasets, and robust feature extraction capabilities. The shortcomings of other algorithms may stem from their excessive complexity, high training requirements, or specific assumptions that do not align with the characteristics of the data.

**Biological Experiments and Bioinformatics Analysis for Auxiliary Validation**

From the machine learning prediction results, 20 samples with probabilities greater than 0.7 were selected for CoExp network analysis. After organizing and analyzing the information of 103 associated genes in the CoExp network, it was found that most of the genes in the CoExp network had annotation information reported to be potentially related to heavy metal biological processes. Some genes had insufficient or overly vague annotation information, making literature investigation impossible. A small number of genes in the database were annotated with terms such as "zinc finger" [27] and "metallothionein" [28], which are clearly related to heavy metal biological processes. Detailed information can be found in Xlsx S2 (CoExp Network Node List).

By comparing the 103 associated genes with significantly expressed genes from RNA-seq, it was found that 12 of the associated genes were identified as mercury-responsive genes through differential expression gene (DEG) analysis in the Hg$^0$ experiments on rice leaves [29]. Among these, 2 genes were samples predicted as heavy metal response-related genes in the prediction set, as detailed in Xlsx S3 and Xlsx S4.

Taking Os10g0317900, predicted as a heavy metal response-related gene in the prediction set, as an example, as shown in Figure 5, its associated genes in the network, Os01g0249200 and Os04g0243700, are annotated in the database with evident heavy metal biological roles: "Superoxide dismutase [Mn]" [30] and "Zinc finger" [27], respectively. The gene Os11g0116300 was validated in Hg$^0$ experiments on rice leaves, showing upregulation at 3h, 9h, and 24h, with significant upregulation observed at 9h and 24h [29].

**Figure 5**

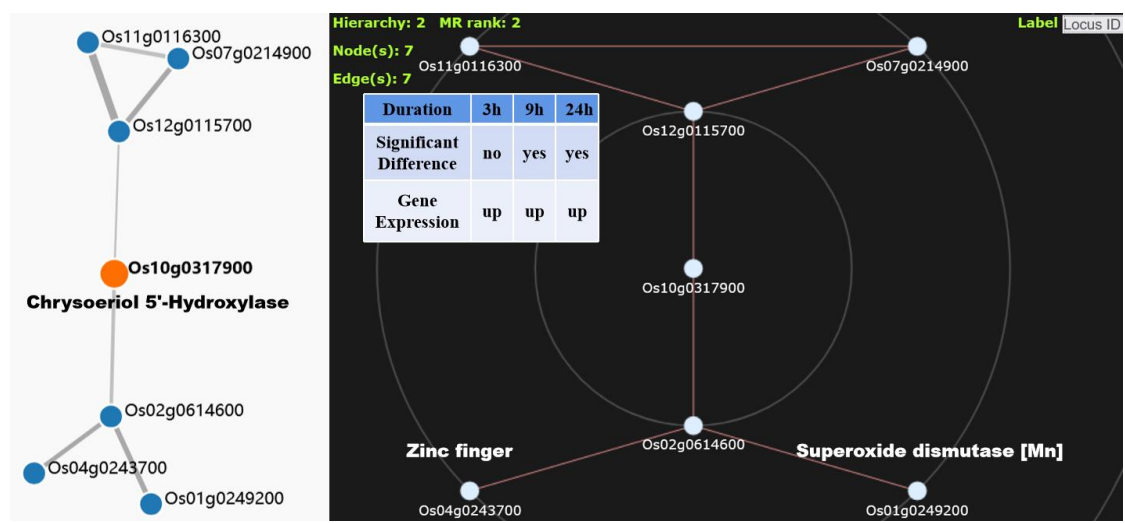

Figure 5. CoExp network schematic diagram, using the gene Os10g0317900 as an example, including existing annotations with clear heavy metal-related tags (Os01g0249200, Os04g0243700) and the gene expression of Os11g0116300 at 3 hours, 9 hours, and 24 hours obtained through RNA-seq and differential gene

expression (DEG) analysis, to study the role of catalase in $Hg^0$ oxidation in rice leaves.

Genes with prediction probabilities greater than 0.7 from the machine learning results exhibited significant heavy metal biological characteristics in the CoExp network. This indicates that the prediction model effectively identifies genes related to heavy metal biological processes. Genes in the CoExp network annotated with heavy metal-related information such as "zinc finger" and "metallothionein", as well as significantly expressed genes from RNA-seq analysis, further support the potential roles of these genes in heavy metal responses.

For example, the role of the Os10g0317900 gene and its associated genes, Os01g0249200 and Os04g0243700, in heavy metal responses is supported by database annotations. RNA-seq validation showed that some samples, such as the Os11g0116300 gene, exhibited significant gene upregulation in $Hg^0$ experiments on rice leaves, particularly at 9h and 24h. This further corroborates the reliability of the machine learning predictions.

The 103 associated genes in the CoExp network, including the 12 differentially expressed genes validated by RNA-seq analysis, provide an important candidate gene pool for studying heavy metal response mechanisms.

In the literature review work, as shown in Xlsx S2 (CoExp Network Node List), by integrating the prediction results with database information and conducting a literature review on the annotated information in the database, it was found that: some annotated information and topics related to biological heavy metal effects have been extensively studied in the literature (e.g., Os05g0112200's "Similar to Zn finger protein (Fragment)"); some annotated information and topics related to biological heavy metal effects have only been reported for the first time in one or two recent papers (e.g., Os07g0683200's "Similar to OsNAC6 protein" [31, 32]), as shown in Figure S4; and some annotated information has not yet been reported in the literature in relation to biological heavy metals (e.g., Os01g0329800's "YT521-B-like protein family protein").

These findings indicate that the integration of machine learning prediction results with database annotation information can effectively identify known, newly discovered, and yet-to-be-studied gene functions. This not only validates the reliability of the machine learning approach but also provides new directions for future research, particularly for those genes that have not yet been reported in the literature, which may become important breakthroughs in future studies.

**Comparison of Biological Experiments, Traditional Bioinformatics Analysis, and New Machine Learning Methods**

In this study, we integrated biological experiments, traditional bioinformatics analysis, and machine learning methods to comprehensively investigate rice genes associated with heavy metal accumulation and conducted an in-depth comparative analysis of the characteristics and applicability of these three approaches.

Biological experiments, as the "gold standard" for validating gene functions, provide direct quantitative and qualitative biological evidence through wet-lab methods such as qRT-PCR [33] and RNA-seq [34], confirming the specific roles of genes in heavy metal responses. However, biological experiments are generally suitable only for small-scale studies due to their time and resource constraints, making them less feasible for large-scale genome-wide gene screening.

Traditional bioinformatics analysis, through methods such as sequence alignment [35], functional annotation [36], and gene co-expression networks [37], provides rich background knowledge and reliable data support for gene function studies. In this study, functional annotation and co-expression network construction helped identify several candidate genes associated with heavy metal responses.

However, traditional bioinformatics methods heavily rely on existing databases and known gene information, making them less effective in predicting unknown genes. Additionally, as the scale of genomic data continues to expand, the efficiency of these methods faces significant challenges.

In comparison, machine learning methods exhibit distinct advantages, as shown in Table 1. By processing large-scale data and extracting features (e.g., k-mer frequency and physicochemical properties), machine learning can rapidly identify new potential functional genes.

In this study, candidate genes identified through machine learning models were validated by RNA-seq and functional annotation analyses. For instance, the heavy metal response function of the Os04g0538000 gene was supported by differential expression gene (DEG) analysis from RNA-seq. Additionally, machine learning methods can construct gene regulatory networks and uncover molecular mechanisms related to heavy metal responses, providing valuable references for subsequent experimental validations.

However, the biological interpretability of machine learning predictions remains limited [38] and requires supplementation with traditional bioinformatics tools and experimental validations.

**Table 1. Comparison of the Characteristics of Biological Experiments, Traditional Bioinformatics Analysis, and New Machine Learning Methods.**

| Characteristics | Biological Experiments | Traditional Bioinformatics | New Machine Learning Methods |
|---|---|---|---|
| Research Scale | Small scale (single-gene validation) | Medium scale (gene set functional analysis) | Large scale (whole-genome prediction) |
| Accuracy | High (but time-consuming and labor-intensive) | Moderate (depends on database accuracy) | High (requires experimental validation) |
| Applicability | Limited by experimental conditions | Database functional annotation analysis | Whole-genome scanning and novel gene prediction |
| Biological Interpretability | Direct and reliable | Strong (based on annotation information) | Weak (requires integration with biological background analysis) |
| Cost and Efficiency | High cost, low efficiency | Moderate cost, relatively high efficiency | Low cost, high efficiency |

In summary, biological experiments, traditional bioinformatics analysis, and new machine learning methods each have their advantages and complement one another in the study of rice heavy metal accumulation genes. Biological experiments are suitable for small-scale gene function validation, traditional bioinformatics analysis plays an important role in gene annotation and medium-scale regulatory network analysis, while machine learning methods demonstrate remarkable efficiency and predictive capability in large-scale whole-genome gene screening.

This study, by integrating the three approaches and using rice heavy metal biological processes as an example, highlights the reliability of machine learning-based genomic linguistics (gene sequence feature learning). It provides a solid foundation for further gene function research and the development of rice varieties resistant to heavy metal stress.

**Significances**

A successful implementation of machine learning-based genomic linguistics (gene sequence feature learning) was achieved, introducing a novel predictive method for studying rice heavy metal response genes. This method is cost-effective, easy to operate, stable in performance, highly reproducible, efficient, fast, high-throughput, and capable of deeper information mining. The approach is highly versatile, making it applicable not only to studying heavy metal responses in rice but also to investigating various gene functions across different species.

Compared to traditional bioinformatics analysis, machine learning offers several advantages. On one hand, it demonstrates stronger information mining and predictive capabilities, evolving gene function prediction methods from traditional bioinformatics tools primarily reliant on sequence alignment algorithms into artificial intelligence models with the ability to learn and understand gene sequences. On the other hand, machine learning overcomes the analytical limitations of traditional tools by shifting gene analysis from the macroscopic level of genome network analysis to the microscopic level of individual gene sequence analysis.

**Current Limitations and Future Prospects**

Since the current rice gene databases contain limited information, the insights that can be mined at present are still relatively constrained. However, the improvement in gene research efficiency achieved through the methods proposed in this study holds the promise of uncovering deeper gene information in the future. This could enable the artificial specific regulation of rice heavy metal responses through genetic engineering and may also be applied to gene function studies in other species.

With comprehensive research on gene information in the future, there is further potential to explore the development of AICG engineering for genomic linguistics.

**Data Availability**

After the paper is published, we will make all raw data, original code, code execution processes, and results from this project publicly available on the Kaggle platform. For those interested, please follow the Kaggle account: Ricky Young (www.kaggle.com/rickyyoung4364). The core information of this project, including essential data and key code, has been concisely organized in the Supporting Information (SI) (DOI: [10.13140/RG.2.2.25798.79685](10.13140/RG.2.2.25798.79685)).

**Supporting Information**

- Describes the process of gene prediction using a convolutional neural network-random forest hybrid model by extracting physicochemical properties and k-mer frequency features from gene sequences, including data collection, feature engineering, model training and evaluation, and validates the model's effectiveness through experiments. Contains the performance results of various models in terms of precision, recall, and F1-score. Additionally, it details the experimental design, RNA extraction and sequencing, differentially expressed gene analysis, and the methods and steps for co-expression network analysis. (PDF)
- Includes core data files in CSV, XLSX, and FASTA formats, as well as a ZIP package containing the core code in different notebooks. (ZIP)

**References**

[1] Chunyapuk, K., Phitchan, S., Nunticha, L., & Supalak, K. (2021) Heavy Metals and Probabilistic Risk Assessment Via Rice Consumption in Thailand, Food Chemistry, 334: 127402-127402.

[2] Estefani Yaquelin, H., Isabel, A., Ana Karina, A., Alfredo, C., & Jose Pedraza, C. (2022) Renal Damage Induced by Cadmium and Its Possible Therapy by Mitochondrial Transplantation, Chemico-Biological Interactions, 361: 109961.

[3] Qudsia, R., Kanwal, R., & Muhammad Sajid Hamid, A. (2021) Heavy Metals and Neurological Disorders: From Exposure to Preventive Interventions, Emerging Contaminants and Associated Treatment TechnologiesEnvironmental Contaminants and Neurological Disorders: 69-87.

[4] Sandeep Kumar, A., & Kavindra Kumar, K. (2019) Mechanistic Effect of Heavy Metals in Neurological Disorder and Brain Cancer, Networking of Mutagens in Environmental Toxicology Environmental Science and Engineering: 25-47.

[5] Colin, R., Noam, S., Adam, R., Katherine, L., Sharan, N., Michael, M., Yanqi, Z., Wei, L., & Peter J., L. (2020) Exploring the Limits of Transfer Learning with a Unified Text-to-Text Transformer, Journal of Machine Learning Research, 21.140: 1-67.

[6] Ming, Z., Nan, D., Shujie, L., & Heung-Yeung, S. (2020) Progress in Neural NLP: Modeling, Learning, and Reasoning, Engineering, 6.3: 275-290.

[7] Qi, F., Debiao, H., Zhe, L., Huaqun, W., & Kim-Kwang Raymond, C. (2020) SecureNLP: A System for Multi-Party Privacy-Preserving Natural Language Processing, IEEE Transactions on Information Forensics and Security, 15: 3709-3721.

[8] Wang, Y. (2018) Intelligent Customer Service System Design Based On Natural Language Processing, International Conference on Electrical and Electronics Engineering: 374-379.

[9] Richard A, Y. (2000) Biomedical Discovery with DNA Arrays, Cell, 102.1.0: 9-15.

[10] Jia-Hong, W., Ling-Feng, Z., Hua-Feng, W., Yue-Ting, W., Kui-Kui, J., Xiang-Ming, M., Zi-Ying, Z., Kai-Tai, Y., Qing-Shan, G., Dan, G., & Zhong-Xi, H. (2019) GenCLiP 3: Mining Human Genes' Functions and Regulatory Networks from PubMed Based on Co-Occurrences and Natural Language Processing., Bioinformatics, 36.6: 1973-1975.

[11] María Katherine, M., & Edward S., B. (2019) A K-Mer Grammar Analysis to Uncover Maize Regulatory Architecture., BMC plant biology, 19.1

[12] Adam, V., Petr, B., Lubomir, A., Petr, S., Ivana, L., & Lucie, H. (2014) Genome Size and Genomic GC Content Evolution in the Miniature Genome-Sized Family Lentibulariaceae., New Phytologist, 203 1: 22-8.

[13] Tomas, P., & Virginijus, M. (2017) Comparison of Naive Bayes, Random Forest, Decision Tree, Support Vector Machines, and Logistic Regression Classifiers for Text Reviews Classification., Baltic Journal of Modern Computing, 5.

[14] Jiuxiang, G., Zhenhua, W., Jason, K., Lianyang, M., Amir, S., Bing, S., Ting, L., Xingxing, W., Gang, W., Jianfei, C., & Tsuhan, C. (2018) Recent Advances in Convolutional Neural Networks, Pattern Recognition, 77: 354-377.

[15] Tawseef Ahmed, T., & Rameez, Y. (2023) Deep Learning for Bioinformatics, Applications of Machine Learning and Deep Learning on Biological Data: 181-196.

[16] Gang, L., & Jiabao, G. (2019) Bidirectional LSTM with Attention Mechanism and Convolutional Layer for Text Classification, Neurocomputing, 337: 325-338.

[17] Li, D., & Dong, Y. (2014) Deep Learning: Methods and Applications, Foundations and Trends in Signal Processing, 7.3-4: 197-387.

[18] Nir, F., Dan, G., & Moises, G. (1997) Bayesian Network Classifiers., Machine Learning, 29.2: 131-163.

[19] NV, C., KW, B., LO, H., & WP, K. (2002) SMOTE: Synthetic Minority Over-Sampling Technique, Journal of Artificial Intelligence Research, 16: 321-357.

[20] S., S., V., M., Y., S., & J., S. (2021) Rule Precision Index Classifier: an Associative Classifier with a Novel Pruning Measure for Intrusion Detection, Personal and Ubiquitous Computing, 27.3: 1395-1403.

[21] Davide, C., & Giuseppe, J. (2020) The Advantages of the Matthews Correlation Coefficient (MCC) over F1 Score and Accuracy in Binary Classification Evaluation, BMC Genomics, 21.1

[22] Wencheng, H., Hongyi, L., Yue, Z., Rongwei, M., Chuangui, T., Wei, X., & Bin, S. (2021) Railway Dangerous Goods Transportation System Risk Identification: Comparisons among SVM, PSO-SVM, GA-SVM and GS-SVM, Applied Soft Computing, 109: 107541-107541.

[23] Liudmila, P., Gleb, G., Aleksandr, V., Anna Veronika, D., & Andrey, G. (2018) CatBoost: Unbiased Boosting with Categorical Features., Conference on Neural Information Processing Systems, 31: 6639-6649.

[24] Karsten, R., Latha, P., Joaquin, Z., Bernhard, S., Thomas, B., & Peter, G. (2022) Towards Total Recall in Industrial Anomaly Detection, Computing Research Repository

[25] Leo, B. (2001) Random Forests, Machine Learning, 45: 5-32.

[26] Swalpa Kumar, R., Gopal, K., Shiv Ram, D., & Bidyut B., C. (2019) HybridSN: Exploring 3-D-2-D CNN Feature Hierarchy for Hyperspectral Image Classification, IEEE Geoscience and Remote Sensing Letters, 17.2: 277-281.

[27] Bogdan, G., Carmen, G., Stelian, D., Anca, B., & Simona, P. (2023) Heavy Metals Acting As Endocrine Disrupters, Scientific Papers Animal Science and Biotechnologies, 44.2: 89-89.

[28] Brian P., M., Jan A., M., Douglas D., R., C. , G., P. H., M., Didier G., S., & Carl E., B. (2013) PHYTOCHELATINS AND METALLOTHIONEINS : Roles in Heavy Metal Detoxification and Homeostasis, semanticscholar.

[29] Weijun, T., Jianxu, W., Yi, M., Christopher W. N., A., & Xinbin, F. (2025) Novel Insights into Hg0 Oxidation in Rice Leaf: Catalase Functions and Transcriptome Responses, ENVIRONMENTAL SCIENCE & TECHNOLOGY, 59.1: 478-488.

[30] Aping, N., Wan-Ping, B., Shuang-Long, F., Shi-Ya, P., Xing-Yi, W., Yi-Fan, Y., Li-Yan, S., & De-Sheng, P. (2020) Role of Manganese Superoxide Dismutase (Mn-Sod) Against Cr(III)-induced Toxicity in Bacteria., Journal of Hazardous Materials, 403: 123604-123604.

[31] Dong-Keun, L., Pil Joong, C., Jin Seo, J., Geupil, J., Seung Woon, B., Harin, J.,

Youn Shic, K., Sun-Hwa, H., Yang Do, C., & Ju-Kon, K. (2017) The Rice OsNAC6 Transcription Factor Orchestrates Multiple Molecular Mechanisms Involving Root Structural Adaptions and Nicotianamine Biosynthesis for Drought Tolerance, Plant biotechnology journal, 15: 754-764.

[32] S. K., P., E., P., S., P., A., P., L., B., S. R., D., & H., P. (2020) Genetic Regulation of Homeostasis, Uptake, Bio-Fortification and Efficiency Enhancement of Iron in Rice, Environmental and Experimental Botany, 177: 104066-104066.

[33] Xinjin, L., Jiangpeng, F., Qiuhan, Z., Dong, G., Lu, Z., Tao, S., Wenjia, H., Ming, G., Xin, W., Zhixiang, H., Yong, X., Guozhong, C., Yu, C., & Ke, L. (2020) Analytical Comparisons Of Sars-Cov-2 Detection By Qrt-Pcr And Ddpcr With Multiple Primer/Probe Sets, Emerging microbes & infections, 9.1: 1175-1179.

[34] Yuqing, Z., Giovanni, P., & W. Evan, J. (2020) ComBat-Seq: Batch Effect Adjustment for RNA-Seq Count Data, NAR genomics and bioinformatics, 2.3

[35] Tsukasa, N., Kazunori D., Y., Kentaro, T., & Kazutaka, K. (2018) Parallelization of MAFFT for Large-Scale Multiple Sequence Alignments, Bioinformatics, 34.14: 2490-2492.

[36] Brad T., S., Ming, H., Ju, Q., Xiaoli, J., Michael W., B., H. Clifford, L., Tomozumi, I., & Weizhong, C. (2022) DAVID: a Web Server for Functional Enrichment Analysis and Functional Annotation of Gene Lists (2021 Update), Nucleic Acids Research, 50.W1: W216-W221.

[37] Rui, Z., Yong, M., Xiaoyi, H., Ying, C., Xiaolong, H., Ping, W., Qi, C., Chi-Tang, H., Xiaochun, W., Youhua, Z., & Shihua, Z. (2020) TeaCoN: a Database of Gene Co-Expression Network for Tea Plant (camellia Sinensis)., BMC Genomics, 21

[38] Marco Tulio, R., Sameer, S., & Carlos, G. (2016) "Why Should I Trust You?" Explaining the Predictions of Any Classifier, North American Chapter of the Association for Computational Linguistics: 1135-1144.